%% file: Zika-TR-03-17.tex
\documentclass{llncs}
 
\input{preamble}

\title{Recruiting from the network: discovering Twitter users who can help combat Zika epidemics}

\author{Paolo Missier \inst{1} \and Callum McClean \inst{1}  \and Jonathan Carlton \inst{1} \and Diego Cedrim \inst{2} \and Leonardo Silva \inst{2} \and Alessandro Garcia \inst{2} \and Alexandre Plastino \inst{3}   \and Alexander Romanovsky \inst{1}}

\institute{School of Computing Science, Newcastle University, UK
\and PUC-Rio, Rio de Janeiro, Brasil 
\and Universidad Federal Fluminense, Niter\`{o}i, Brasil
}

\begin{document}
\maketitle

\begin{abstract}
Tropical diseases like \textit{Chikungunya} and \textit{Zika} have come to prominence in recent years as the cause of serious, long-lasting, population-wide health problems.
In large countries like Brasil, traditional disease prevention programs led by health authorities have not been particularly effective.
We explore the hypothesis that monitoring and analysis of social media content streams may effectively complement such efforts. Specifically, we aim to identify selected members of the public who are likely to be sensitive to virus combat initiatives that are organised in local communities.
Focusing on Twitter and on the topic of Zika, our approach involves (i) training a classifier to select topic-relevant tweets from the Twitter feed, and (ii) discovering the top users who are actively posting relevant content about the topic. We may then recommend these users as the prime candidates for direct engagement within their community.
In this short paper we describe our analytical approach and prototype architecture, discuss the challenges of dealing with noisy and sparse signal, and present encouraging preliminary results.
\end{abstract}

\input{TR-Introduction}

\input{TR-Offline}

\input{TR-Online}

\input{TR-Results}

\input{Zika-TR-03-17-static.bbl}

\end{document}

%% file: preamble.tex
\usepackage[utf8]{inputenc}
\usepackage{eurosym}

\usepackage{graphicx}
\usepackage{float}

\usepackage[fleqn]{amsmath}
\usepackage{tabularx}
\usepackage{tabulary}
\usepackage{mathpartir}
\usepackage{amssymb}
\usepackage{subfiles}
\usepackage{soul}
\usepackage{booktabs}
\usepackage{subfigure}

\usepackage{color}
\usepackage{xcolor}
\usepackage{framed}
\definecolor{shadecolor}{rgb}{0.9,0.9,0.9}
\definecolor{Orange}{rgb}{1,0.5,0}

\usepackage{mathtools}
\usepackage{cite}
\usepackage{epsfig}
\usepackage{epstopdf}
\usepackage{algorithm}
\usepackage{algorithmic}
\usepackage{caption}
\usepackage[pdftex]{hyperref}
\hypersetup{%
pdfauthor={blind submission}, %
bookmarksnumbered, %
pdfstartview={c}, %
colorlinks,%
citecolor=black, %
filecolor=black, %
linkcolor=black, %
urlcolor=black}

\floatname{algorithm}{Procedure}

%% file: TR-Introduction.tex
\section{Introduction}

Mosquito-borne disease epidemics are becoming more frequent and heterogeneous in tropical and subtropical areas around the world. Indeed, we witness the rapid rise to prominence of the \textit{Chikungunya} and \textit{Zika} viruses \cite{MilesHirschler2016}. These viruses together with the \textit{Dengue} virus are responsible for thousands of deaths every year \cite{Denguecenter2015}, as well as for long-lasting health problems, especially to children. To make the matter worse, there is a potential relation between Zika virus infection and birth defects \cite{Rasmussen2016}. In Brazil, in particular, the regional focus of our research, disease prevention programs led by health government authorities have not been particularly effective. For instance, Brazilian Health System requires that health agents report each Zika case; however, it takes several days to process and publish such information. 


Due to the inefficiency of health government programs, no one surprises that the Brazilian population has been so engaged in sharing mosquito-related content on social channels. In fact, the population has shared a variety of types of information, including complaints about personal health, dissemination of public news, but also, importantly, details about the discovery of mosquito breeding sites in public locations. In spite of the volume of mosquito-content, real-time social media is potentially a much faster vehicle for information than traditional channels. Furthermore, together with the shared content, some users stand out for the quality and relevance of their contribution to the social media. These users are namely \textit{social sensors}. The term \textit{social sensors} has been used in similar contexts \cite{Sakaki2010}, to denote portions of the online population that spontaneously contribute with information on social media channels, which is relevant to a particular topic.

As social sensors are influential references on social media, this short paper presents an initial investigation into the kind of social sensor signals that can be effectively detected from real-time social media streams. The goal is to rank users who can act as social sensors. Our approach to rank users is based on the classification of relevant tweets. First, we classify tweets automatically based on their content. Classification aims at filtering out relevant tweets from irrelevant ones. Second, we apply an adaptation of the TwitterRank algorithm to rank users who authored the relevant tweets. Additionally, this paper investigates how social sensors can be exploited to complement and support institutional disease combat efforts. Specifically, we investigate the hypothesis that \textit{real time, short content} social media websites such as Twitter, Instagram, etc., when appropriately analysed, are strong allies on the combat and prevention programs. That is, these networks can be exploited to engage the population on health programs by selecting members of online communities (social sensors) to contribute to health vigilance in their \textit{local} communities. Ultimately, we aim to support health authorities, as they need to engage the population to embrace the combat and prevention programs. This support happens when we rank influential users (social sensors) who can engage communities' members.

Our solution to reveal and rank social sensors is integrated to our VazaZika portal \footnote{Available at http://vazadengue.inf.puc-rio.br/}. VazaZika works as an entomological surveillance system in order to combat the mosquito that transmits Zika, Chikungunya, and Dengue. The portal and a mobile app allow users to report and visualize occurrences of the mosquito or cases of sick people. VazaZika is integrated to social medias in order to reveal social sensors in such medias. Our solution plays an important role to popularize the surveillance system and the engagement programs provided by the VazaZika portal.

\subsection{Overview of the approach}  \label{sec:overview}

\begin{figure}
\centering
   \includegraphics[width=\linewidth]{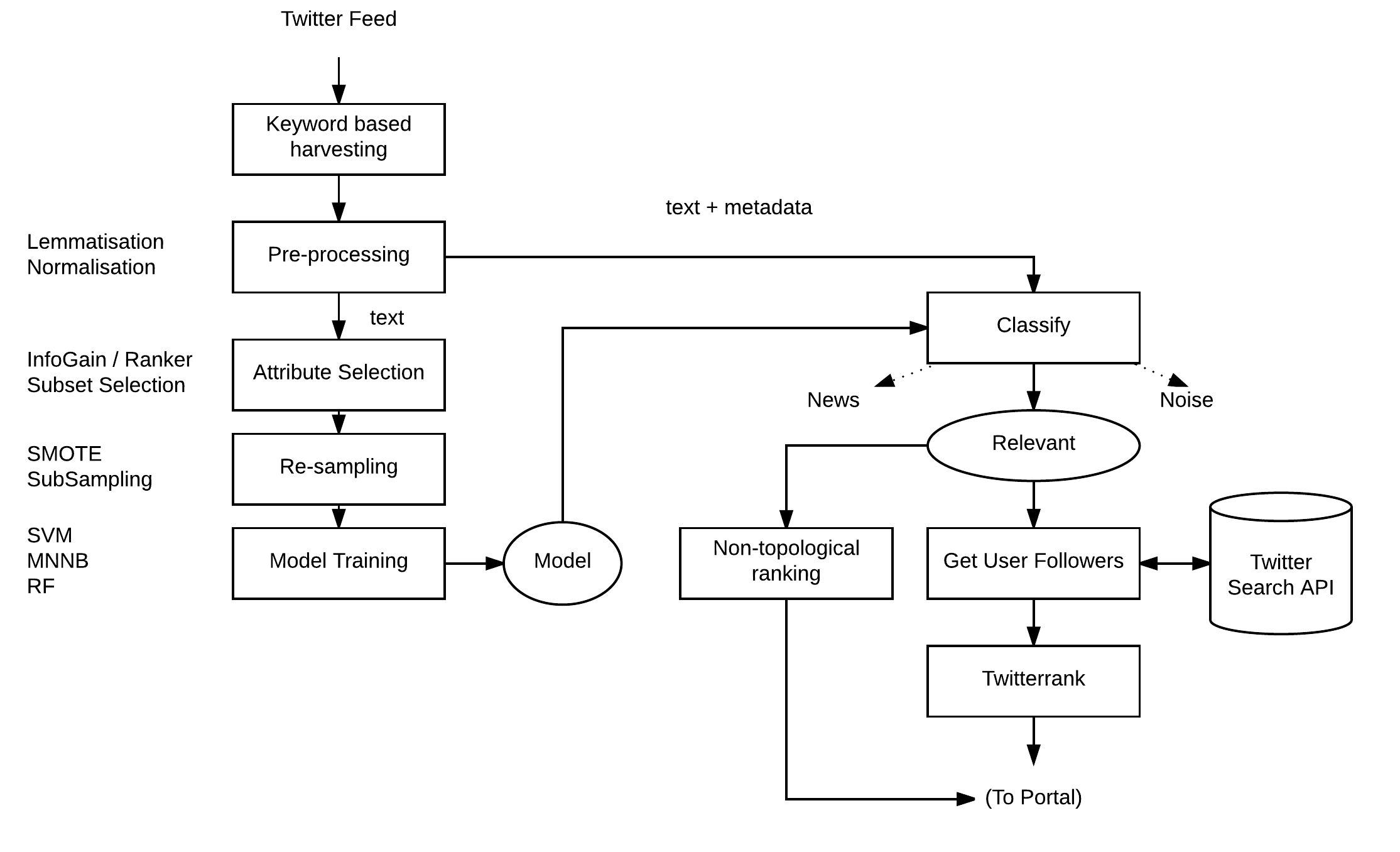}
   \caption{Dataflow diagram for content classification and user ranking}
   \label{fig:dataflow}
\end{figure}

Our approach combines content-based automated classification of tweets, aimed at isolating the sparse relevant signal out of generally noisy chatter about Zika, followed by a ranking of the users who author such relevant content. 
This is summarised in the dataflow diagrams of Fig. \ref{fig:dataflow}.

Initially, in the \textit{offline phase} (left in the figure), a classifier is trained on a collection of manually annotated tweets.
The classifier aims at segregating the \textit{target tweets} that are indicative of user interest in aspects of the Zika problem, as opposed to news feeds, e.g. those originating from news agencies, as well as background noise.
The main challenges in achieving good classification performance is the high levels of noise found in the filtered harvest. This is mainly due to the idiosyncratic use of critical keywords, such as \textit{Zika} itself, which in Brasil happens to be used as a common slang word completely out of the context of discourse about the virus or the disease. 
In the wild, the target tweets are  less than 10\% of a typical harvest.

In the \textit{online phase} (right side of the figure), tweets are continuously harvested from the raw twitter feed, using a set of filtering keywords that we have chosen to provide high recall relative to the set of target tweets. 
We denote as \textit{candidate users} the authors of all tweets that are classed as \texttt{Relevant}.
These are ranked using a variation of the TwitterRank algorithm \cite{weng2010twitterrank}, which we modify to operate on a single topic.
For this, the users connections in the Twitter social graph are retrieved (specifically, the set of users' followers) and used to rank the candidate users according to their relative relevance. 
Ideally, this approach provides a set of top-k target users, which is continuously updated as the live feed is tracked over time.

\subsection{Challenges and Contributions}

In this short paper we present our technical approach, including our experimental selection of a suitable classification approach, our single-topic adaptation to TwitterRank, and some early results.
While classification accuracy is acceptably good (84.1\%, F-measure = .84), we find that acquiring a sufficient number of relevant tweets per user requires a long harvest time.
On a 3-months batch in 2016, we have identified about 13,000 relevant tweets, with most users only contributing one single relevant tweet.
 This sparsity of users suggests that at this time scale it is difficult to apply any ranking criterion, and that TwitterRank is both ineffective and inefficient. 
 Indeed, TwitterRank assumes knowledge of the social graph neighbourhood for each candidate user, and requires that meaningful social connections exist within those neighbourhoods.  
Thus,  it is an inefficient approach because it requires retrieving all followers for a large number of single-contribution candidate users; and it is ineffective because the vast majority of these followers will not be candidate users themselves, which means they will not contribute to the ranking of other users.
In reaction to these observations, in Sec. \ref{sec:results} we propose alternative, and simpler, ranking criteria that do not rely upon the topological properties of the social graph around the users, and compare those with the TwitterRank  top-k users.
Note that, as in the original TwitterRank research \cite{weng2010twitterrank}, no ground truth, i.e., explicit knowledge of these top users, is available for evaluation, as our content harvesting was performed purely ``in the wild''.
Thus, our discussion on the results is necessarily based on a comparison of relative merits of these metrics.

\subsection{Related work}

Similar to our work, \cite{Yamaguchi2010} propose a method to rank Twitter users using a variation of graph analysis called TURank. The authors perform link structure analysis on the user-tweet graph they introduce consisting of tweets and users as nodes, and follow and retweet relationships as edges. While we do not consider the retweet relationships that users form between each other, as our ranking phase does not allow for it, we do evaluate the tweets that users are posting.

In \cite{Chen2014} they propose a clustering algorithm to partition influential users into five categories; fan, disseminator, expert, celebrity, and others. Their work relates to ours as the clustering of users into the five influence role categories can be related to our topic-specific communities, however, their communities are formed of users that are experts, for example, on multiple topics. The authors introduce a limitation in their work; they pick the top 10 users within a topic and crawl each of their followers, making an assumption that those followers also have an interest in that specific topic. Our approach tests the validity of the followers of a relevant user by only considering followers that are relevant themselves.

In our previous work \cite{Missier2016a}, we used topic modelling similar to that shown in \cite{Surian2016}, however, we focused on pre-defined classes of interest specifically related to Zika epidemics. They use community detection (Louvain modularity) and the encoding of random walks to detect community structures within the topics they previously defined. However, in our experiments, we found that the data, tweets and users, are too sparse to form communities, thus other approaches are required, i.e. ranking.

An alternative approach to finding authoritative users and ranking them is presented in \cite{Wei2016} by Wei \textit{et al}. They use a combination of Twitter lists (a grouping of followers per a criterion), the follower graph and the users profile information to produce a global authority score for each user in their data set. In this paper, we do not use the lists nor the user profile but it demonstrates that Twitter is a great resource, with many attributes, that can be utilised to produce results similar to ours.

A heuristic-based approach for automated identification of expertise on Twitter is presented in \cite{Horne2016} and is based on the premise that experts will use Twitter differently to that of non-experts. They find that experts tend to receive information from many friends, filter and distil it, and that experts tend to be old, in relation to the length of time passed since the creation of their accounts, Twitter users. Our work differs as we aim at seeking out users who stand out not because of their expertise but because of their demonstrated interest in engaging with a specific topic.

%% file: TR-Offline.tex
\section{Twitter relevance model training}  \label{sec:offline}
%
%
%
%

We mentioned in the introduction that user ranking requires first of all the capability to identify  with high precision the few tweets that are relevant  to the Zika topic, amongst a large amount of Twitter noise.
For this, we tuned a harvester on a set of relevant keywords, and then trained a supervised classifier on an initial set of about 10,000 manually annotated tweets, collected over multiple time windows, between Sept and Dec. 2016.
As mentioned earlier, however, the need to use the keyword \textit{Zika} makes the harvesting and initial filtering difficult and results in a particularly noisy dataset.
Fine-tuning of data pre-processing and model training was therefore required.
We discuss these issues in the rest of this section.

\subsection{Selecting Twitter feed harvesting keywords}  \label{sec:keywords}

The first task, Twitter harvesting (top of Fig.\ref{fig:dataflow}) provides content both for manual annotation and model training, as well as for classification and then user ranking.
High recall is important in the initial filtering, as the relevant tweets we seek to isolate are no more than about 10\% of the feed. At the same time filtering out clearly irrelevant content is required for reducing the noise prior to classification.
The choice of keywords to harvest from the live Tweeter feed\footnote{For this we used the Twitter stream API through the Twitter4j library.} is therefore critical to striking this balance.

Filtering keywords were selected in two steps, following an approach similar to that suggested in \cite{Nagarajan2009}. Firstly, a short list of \textit{seed} keywords was \textit{bootstrapped} from sample tweets content using manual, expert inspection, and borrowing from our earlier work \cite{Missier2016a}.
These are the top 8 keywords: 
\texttt{dengue}, \texttt{combateadengue}, \texttt{focodengue}, \texttt{todoscontradengue}, \texttt{aedeseagypti}, \texttt{zika}, \texttt{chikungunya}, \texttt{virus}.

Those keywords were then used to harvest an initial corpus of tweets, whose terms were then ranked according to their TF-IDF score relative to the corpus. The top 10 of those were added to the initial seed set, after removing common stopwords and those words that experts deemed to be out of context. 
The resulting additional terms, listed here below, were used together with the seed terms above, as filtering keywords on the Tweet stream API, both for harvesting the training set, and for ongoing harvesting for continuous user ranking:
\texttt{microcefalia}, \texttt{transmitido},
\texttt{epidemia},
\texttt{transmissao},
\texttt{doenca},
\texttt{eagypti},
\texttt{doencas},
\texttt{gestantes},
\texttt{infeccao},
\texttt{mosquitos}.

\subsection{Learning a relevance model}

We aimed to learn a classifier that effectively provides an operational definition of \textit{relevance} of tweets in the context of our topic.
In our previous work on detecting Dengue-related tweets \cite{Missier2016a} we used four target classes with the same purpose: \texttt{Mosquito-focus}, \texttt{Sickness}, \texttt{News}, and \texttt{Joke}, representing (i) content that is strictly relevant to the topic, (ii) content that describes symptoms by affected people, (iii) content from news agencies or that echoes news from agencies, and (iv) content from people who make mostly sarcastic or humorous remarks about Dengue, respectively.
In that setting, both \texttt{Mosquito-focus} and \texttt{Sickness} tweets would be considered relevant.

For Zika-related content, we focused initially on two ``relevance'' classes, namely information \texttt{provider} and \texttt{receiver}, with a view to engage two groups of users: those who are shown to volunteer information about possible infestation locations, the \texttt{providers}, as well as those who may need assistance because they talk about their experience being infected, i.e., the \texttt{receivers}.
However, the more noisy nature of Zika content relative to the Dengue content, along with the scarcity of instances in each of the two relevant classes, contributed to poor accuracy in our early experiments, suggesting that merging the two classes might be beneficial.
In this work, we therefore only use three classes: \texttt{Relevant}, \texttt{News}, and \texttt{Noise}.

In the work just cited, we contrasted a traditional supervised learning approach (a Naive Bayes model) with unsupervised topic modelling, using variations of the LDA algorithm \cite{Blei:2003:LDA:944919.944937}, which has proven popular in recent research~\cite{Morstatter2013,Rosa2011}.
We concluded that LDA under-performs when topics are pre-selected and topic modelling is expected to discover ``sub-topics'', and that a relatively small annotation effort (2,000 tweets at the time) was sufficient (.83 F-measure across the classes).

Having noted earlier that high recall \textit{Zika} Twitter harvests are going to be more noisy than the more specific \textit{Dengue} tweets, in this work we have focused solely on supervised classification, using 10,000 labelled examples.
We experimented with a number of supervised classification models as well as multiple data preparation steps, illustrated in Fig. \ref{fig:dataflow}, left side, all implemented using the Weka toolkit.

The final configuration, described below, is the result of exploration over a space of available alternatives and parameter settings at each step of the data preparation pipeline.
This includes
(i) representing tweets using bag-of-words and a choice of N-grams (N=1,2,3); (ii) attribute selection using Ranking with Information Gain vs Subset Selection; and (iii) whether to rebalance class distribution in the training set, i.e. using class over- and sub-sampling  (note that Attribute Selection methods, namely Ranking with Information Gain and Subset Selection, did not improve performance and are therefore not discusssed further).

For the initial \textit{text normalisation} we used POS tagging and lemmatisation\footnote{We used the tagger from Apache OpenNLP 1.5 series (\url{http://opennlp.sourceforge.net/models-1.5/}), and the LemPORT Lemmatizer customised for  Portuguese language vocabulary.}, also removing common regional ``twitter lingo'' abbreviations, as well as all emoticons and non-verbal forms of expressions.
While those are crucial to understanding the \textit{sentiment} expressed in a tweet, we found that they are not good class predictors.
Links, images, numbers, and idiomatic expressions were also replaced by conventional terms (\textit{url}, \textit{image}, \textit{funny},...).\footnote{Note that these steps are the same as described in \cite{Missier2016a}).}

In searching for a suitable combination, we then heuristically reduced the space of possible configurations by first establishing a baseline classifier performance, for three popular classification models that have proved effective for short text classification\cite{DBLP:conf/ecai/CarvalhoP16}: Support Vector Machines (SVM), Multinomial Naive Bayes (MNNB), and Random Forest (RF).
This includes a choice of N-grams, but no attribute selection and no class re-sampling.
Table \ref{tab:baseline-accuracy} reports the overall accuracy and F-measure (in parenthesis) for each of these classifiers.
Based on these early results, We ruled out SVM, which performed substantially more poorly, and focused solely on MNNB and RF.

\begin{table}[t]
\centering
\begin{tabular}{c c c c } 
& \textbf{1-grams} & \textbf{1+2-grams} & \textbf{1+2+3-grams} \\ \midrule
\textbf{SVM} &  73.96 (0.68) & 73.97 (0.70) & 74.01 (0.70) \\  
\textbf{MNNB}  &  81.21 (0.81)& 81.74 (0.82) & 81.81 (0.82)\\ 
\textbf{RF} &  81.1 (0.80) & 80.65 (0.80)& 79.97 (0.79)\\ \bottomrule
\end{tabular}
 \caption{\scriptsize Baseline classifier accuracy}
  \label{tab:baseline-accuracy}
\end{table}





\paragraph{Class rebalancing.} As mentioned above, one of the main classification challenges is the relative scarcity of \texttt{Relevant} tweets in the Twitter feed for user ranking.
This imbalance in the minority class is naturally reflected in the class proportions observed in the training set: 50.6\% \texttt{News}, 37.3\% \texttt{Noise}, 12.1\% \texttt{Relevant}, and may reduce accuracy.
To address this issue, we experimented with two complementary approaches. 
Firstly, we added an extra 600 annotated examples to the \texttt{Relevant} class.
Secondly, we applied statistical over-sampling to the \texttt{Relevant} class, using the SMOTE algorithm \cite{Chawla2002} to boost the examples from 1,214 to 2,428 (12.1\% to 24.3\%). 

The results, reported in Table \ref{tab:best-performance} for various combinations of N-grams and MNNB vs RF, show that  there is no real advantage in investing extra human annotation effort, as boosting using SMOTE provides equivalent performance.
Note that the results also show that down-sampling the majority class (\texttt{News}) is not as beneficial.

\begin{table}[t]
\centering \scriptsize 
\begin{tabular}{c c c c | c c c } 
& \multicolumn{3}{c}{\textbf{RF}}  & \multicolumn{3}{c}{\textbf{MNNB}} \\ 
 \cline{2-4}   \cline{5-7}
& 1-grams & 1+2-grams & 1+2+3-grams & 1-grams & 1+2-grams & 1+2+3-grams  \\  \midrule
SMOTE over-sampling &  83.5 & 83.1 & \textbf{84.1} & 81.2  & 80.9 & 81.2 \\ 
Sub-sampling (Spread) &  75.8 & 76.3 & 76.1 & 77.5  & 78.9 & 79.95 \\ 
Over- and sub-sampling &  82.5 & 82.7 & 83.6 & 80.6  & 80.0 & 80.95 \\ 
+600 \texttt{Relevant} samples  &  80.8 & 80.5 & 80.4 & 80.5  & 81.0 & 81.2  \\ \bottomrule
\end{tabular}
 \caption{\scriptsize Classifier accuracy for various choices of N-grams and over- and sub-sampling}
  \label{tab:best-performance}
\end{table}


The Table also reports the best overall accuracy figure across all configurations, namely 84.1\%, obtained from a Random Forest learner (using an ensemble of 100 trees), with 1,2,3-grams, no attribute selection, and SMOTE-based boosting.
More in detail, the performance measures of this configuration is: weighted average F-measure=0.84 across the three classes is, with F-measure=0.83 for the \texttt{Relevant} class, and RMSE=0.28.
This is the classifier we used for the online content relevance detection phase in combination with user ranking, described next.

%% file: TR-Online.tex
\section{User ranking}  \label{sec:ranking}

In the next phase of our study, we collect all users that have authored at least one \texttt{Relevant} tweet and experiment with three ranking criteria to select the top-k users.
While we hope these may be ideally suited for engagement by the health authorities on Zika combat campaigns, we have no ground truth about the effective attitude of these users, as our study is conducted entirely \textit{in the wild}.
Thus, we are going to present our results in the form of a comparative analysis across the three types of rankings.
Specfically, we compare our own variation of the TwitterRank algorithm \cite{weng2010twitterrank}, which is based on the social media graph, with non-topological metrics that simply count the fraction of relevant tweets per user within the harvest set and within the whole twitter stream.
Firstly, we describe these metrics.

\subsection{TwitterRank}
In \cite{weng2010twitterrank} a method of assigning a topic-specific rank to the users of Twitter is proposed, called TwitterRank (TR). The approach is an extension of PageRank, however, TR differs as it measures importance by taking both the topical similarity between users and the underlying social network structure into account. They propose the formula below to calculate the topic-specific rank for a user.

       \[ \overrightarrow{TR_t} = \gamma P_t \times \overrightarrow{TR_t} + (1 - \gamma)E_t \]

$\overrightarrow{TR_t}$ is the TR score associated with a user for topic $t$. $P_t$ is the transition probability of a random surfer moving from follower to friend. $E_t$ is the teleportation vector of the random surfer in topic $t$, i.e. how many times a user's tweets have been assigned to topic $t$. $\gamma$ is a variable that controls the probability of teleportation. The lower $\gamma$ is the higher the probability that the random surfer will teleport to users according to $E_t$ and vice versa \cite{weng2010twitterrank}.

\subsection{Adaptation of TwitterRank}
While TR can fit with our work, we found that TR does not contextually translate perfectly and needed adapting slightly in order to work with our data sets. The authors, in \cite{weng2010twitterrank}, use a multi-column matrix to store the rank of a user within their data set, with each column representing a topic and each row a user. We limit this matrix to a single topic, as we're interested in discovering highly ranked users within a topic-relevant virtual community. Furthermore, they propose a topical difference between two users, which isn't applicable in our context so we introduce a new metric; the normalised occurrences for a user: $v_t$.

The transition probability is calculate as shown below; this determines the likelihood that a random user will start at follower $s_i$ and then move to $s_j$.

    \[ PT_t(i,j) = \frac{|\tau_j|}{\sum\limits_{a: s_i  follows  s_a}|\tau_a|} \times sim_t(i, j) \]

This formula, fundamentally, remains the same for us, however, we redefine components of it. To start, $\tau_j$, the number of tweets published by $s_j$, is changed to the number of tweets published by $s_j$ within the topic (rather than overall). $\tau_a$ becomes the number of tweets published by all of $s_i$'s friends, that are within the topic, rather than the sum of tweets published by all of $s_i$'s friends across all topics. Finally, $sim_t(i, j)$ calculates the similarity between $s_i$ and $s_j$ within topic $t$. We changed this to find the absolute different $v_t$ for users $i$ and $j$, rather than the absolute topical difference between users $i$ and $j$. The change is shown below respectively.

	\[ sim_t(i, j) = 1 - |DT'_{it} - DT'_{jt}| \] 
	\[ sim_t(i, j) = 1 - |\upsilon_{it} - \upsilon_{jt}| \] 
	
The final modification that we made was to the teleportation vector for the random user, $E_t$. Originally this described the $t$-th column of the topical difference matrix and is the column-normalised form of the matrix $DT$ such that $||DT''_{.t}||_1 = 1$. This isn't a metric that we use, however, it forms part of the overall TR calculation. Therefore, we instead use the normalised occurrences for a user $i$ in topic $t$.

\subsection{Application of TwitterRank}
We create a Java-based application in order to implement the adaption of TR presented previously. To start, the followers for each user within the data set are collected using a crawler previously developed that queries the Twitter public REST API. Once the followers are collected, we only consider followers of a user if all of those or a subset of followers are also in the data set. This approach iteratively builds topic-specific communities starting with one user and then expanding outwards, potentially linking communities together. We decided to do this as there would be a lot of noise introduced in calculating the TR score as the vast majority of those in the social network would not be relevant to our goal; if all followers are considered, and it reduces a computation overhead discussed in \cite{Bartoletti2016}. Finally, as per the original TR approach, we set the teleportation vector as $\gamma = 0.85$.

\subsection{Non-topological metrics}

For a user $u$ and a set $K$ of keyword, let $T_K$ denote the entire harvest, $T_K(u)$ the number of tweets in $T_K$ that are attributed to $u$,  $R_K(u)$ the number of \texttt{Relevant} tweets in $T_K(u)$, and $T(u)$ the \textit{total} number of tweets posted by $u$ during the harvest period.

We define the \textbf{Topic Focus} per user as $TF(u) = \frac{R_K(u)}{T_K(u)}$. This is the fraction of $u$'s tweets in the harvest, which are \texttt{Relevant}, an indication of how often user $u$ used the keywords $K$ to express relevant content;

We define the \textbf{Overall Focus} per user as $TF(u) = \frac{R_K(u)}{T(u)}$. This is the fraction of $u$'s total tweets in the harvest period, which are \texttt{Relevant}. We take this as an indication of the focus of the user on the topic, considering the user's global interests when posting on Twitter.

%% file: TR-Results.tex
\section{Results}  \label{sec:results}


\subsection{Experimental dataset}

Given a keyword-based harvest from the Twitter feed, we refer to the set of users who have posted at least one \texttt{Relevant} tweet as the \textit{candidate} users.
The \textit{target} users are the top candidate users according to some ranking criteria.
In this Section we report our preliminary findings on characterising candidate users and ranking them to discover target users.

Our experimental dataset consists of a harvest of 278,351 tweets, collected and classified through our online pipeline (Fig.\ref{fig:dataflow}) using the keywords presented in Sec. \ref{sec:keywords} during a period of 4 months (9-12) in 2016.
Using our classifier, we found 15,124 \texttt{Relevant} tweets in this set.

Firstly, we note that the vast majority of those users only produced one single or very few \texttt{Relevant} tweets during the harvest period, as 
shown in Tab.\ref{tab:TUD}.
This means that there are very many candidate users (13,228 in our batch), each producing a very weak signal both in terms of generated content and in terms of their social connections to other candidate users.

\begin{table}
\centering
\begin{tabular}{c c } 
\textbf{\texttt{Relevant} Tweets}	& \textbf{Users count} \\  \midrule
$\geq$ 20	& 2 \\
(10,19)	&  1 \\
(5,9) &	41 \\
4	& 57 \\
3	& 209 \\
2	& 1058 \\
1	& 11860 \\ \bottomrule
\end{tabular}
\caption{Distribution of \texttt{Relevant} tweets per candidate user}
\label{tab:TUD}
\end{table}

To deal with this long tail and to strike a balance between strength of content signal and numerosity of candidate users, we only considered users who posted at least 3 \texttt{Relevant} tweets. 
Out of these 310 users, however, we had to exclude a further 139 whose followers could not be obtained due to privacy settings, leaving 171 candidate users for ranking.
The results presented below concern these users.

Tables \ref{tab:TR}, \ref{tab:TF}, and \ref{tab:OF} show the top 10 users ranked according to each of our three criteria (TwitterRank, Topic Focus, and Overall Focus), respectively.
For each of these users, each table also shows the values for the other two metrics, and the position of that user when ranked according to those metrics.

\begin{table}
\centering \scriptsize 
\begin{tabularx}{\textwidth}{p{2.5cm} | p{2cm} |X |X |X |X | X} 
\textbf{Screenname}	&\textbf{Twitterrank (x100)}&	\textbf{Relevant count}&	\textbf{Overall focus (x100)}&	\textbf{OF Rank}	&\textbf{Topic \newline focus}& 	\textbf{TF Rank} \\  \midrule
FlorzinhaSimoes	& 0.84 & 20	& 14.28 & 3	& 71.428 & 15 \\ 
Lorrayn54837060	& 0.64 & 3	& 0.1708 & 142	& 75& 	14 \\
pelotelefone	& 0.41 & 7	& 6.1947 & 7	& 87.5	& 7 \\
SEIZETHEHEAVEN	& 0.39 & 7	& 0.3693& 65	& 100	& 1\\
macabia	& 0.39 & 3	& 0.44 & 55& 	100 & 5 \\
gushfsc	& 0.37 & 6	& 0.30 & 85	& 60 & 18 \\
tiiancris	& 0.37 & 3	& 0.19 & 128	& 50& 24 \\
scomacinha	& 0.35 & 3	& 0.13 & 164	& 33.33 & 28 \\
sophiaboggiano	& 0.35 & 3	& 0.14 & 160	& 75	& 14 \\
mariabarrozoo	& 0.34 & 3	& 0.11 & 169	& 60	& 19 \\ \bottomrule
\end{tabularx}
\caption{\scriptsize Top 10 TwitterRank candidate users}
\label{tab:TR}
\end{table}

\begin{table}
\centering \scriptsize 
\begin{tabularx}{\textwidth}{p{2.5cm} | p{1cm} |X | p{1cm} | p{1cm} |X |X |X} 
\textbf{Screenname}	&\textbf{Topic Focus}&	\textbf{Relevant count}&	\textbf{All tweets count}&	\textbf{Overall focus (x100)}	&\textbf{OF Rank}& 	\textbf{TR (x100)}& 	\textbf{TR position} \\  \midrule
SEIZETHEHEAVEN	& 100 & 7 & 1895 & 0.3693	& 65 & 0.39 & XX \\ 
LairaMaia		& 100 & 6 & 799  & 0.7509	& 35 & 0.07 & XX \\
llGueto			& 100 & 6 & 1427 & 0.4204	& 58 & 0.07 & XX \\
Giovannacoosta	& 100 & 5 & 960  & 0.5208	& 45 & 0.06 & XX \\
pakito\_lucas	& 100 & 5 & 2149 & 0.2326   & 111 & 0.06 & XX \\
Lorranna\_Castro& 100 & 5 & 1573 & 0.3178	& 84 & 0.06 & XX \\
laricrvlh		& 100 & 5 & 951  & 0.5257	& 43 & 0.06 & XX \\
mauriciooasn	& 100 & 4 & 495  & 0.8080	& 33 & 0.04 & XX \\
masoqmath\_		& 100 & 4 & 2412 & 0.1658	& 145 & 0.04 & XX \\
isaah13\_ferreir& 100 & 4 & 272  & 1.4705	& 19 & 0.04 & XX \\ \bottomrule
\end{tabularx}
\caption{\scriptsize Top 10 Topic Focus candidate users}
\label{tab:TF}
\end{table}

\begin{table}
\centering \scriptsize
\begin{tabularx}{\textwidth}{p{2cm} | p{1.5cm} |p{1.5cm} | p{1cm} | p{1.5cm} |X |X |X |X} 
\textbf{Screenname}	&\textbf{Relevant Count}&	\textbf{Keyword count}&	\textbf{All tweets count}&	\textbf{Overall focus (Rel/All)}	&\textbf{Topic Focus}& 	\textbf{TF rank}& 	\textbf{TR}& \textbf{TR position} \\  \midrule
leilaquintsepe	& 4 & 4 & 19 & 21	& 100 & =4 & 0.04 & 70 \\ 
DCGRodrigues		& 3 & 3 & 18  & 16.6	& 100 & =5 & 0.03 & 169 \\
FlorzinhaSimoes			& 20 & 28 & 140 & 14.2	& 71.4 & 15 & 0.8 & 1 \\
RobelioValle	& 3 & 4 & 31  & 9.6	& 75 & =14 & 0.03 & 156 \\
iaedayana	& 3 & 3 & 37 & 8.1   & 100 & =5 & 0.03 & 125 \\
iPedersoly& 4 & 5 & 51 & 7.8	& 80 & =10 & 0.04 & 81 \\
pelotelefone		& 7 & 8 & 113  & 6.1	& 87.5 & =7 & 0.4 & 3 \\
tacianebielinki	& 6 & 10 & 136  & 4.4	& 60 & =18 & 0.07 & 32 \\
isaldcunha		& 3 & 4 & 98 & 3	& 75 & =15 & 0.03 & 147 \\
onelastovada& 7 & 9 & 285  & 2.4	& 77.7 & 11 & 0.1 & 24 \\ \bottomrule
\end{tabularx}
\caption{\scriptsize  Top 10 Overall Focus candidate users}
\label{tab:OF}
\end{table}

\begin{figure}
\centering
\includegraphics[width=.8\linewidth]{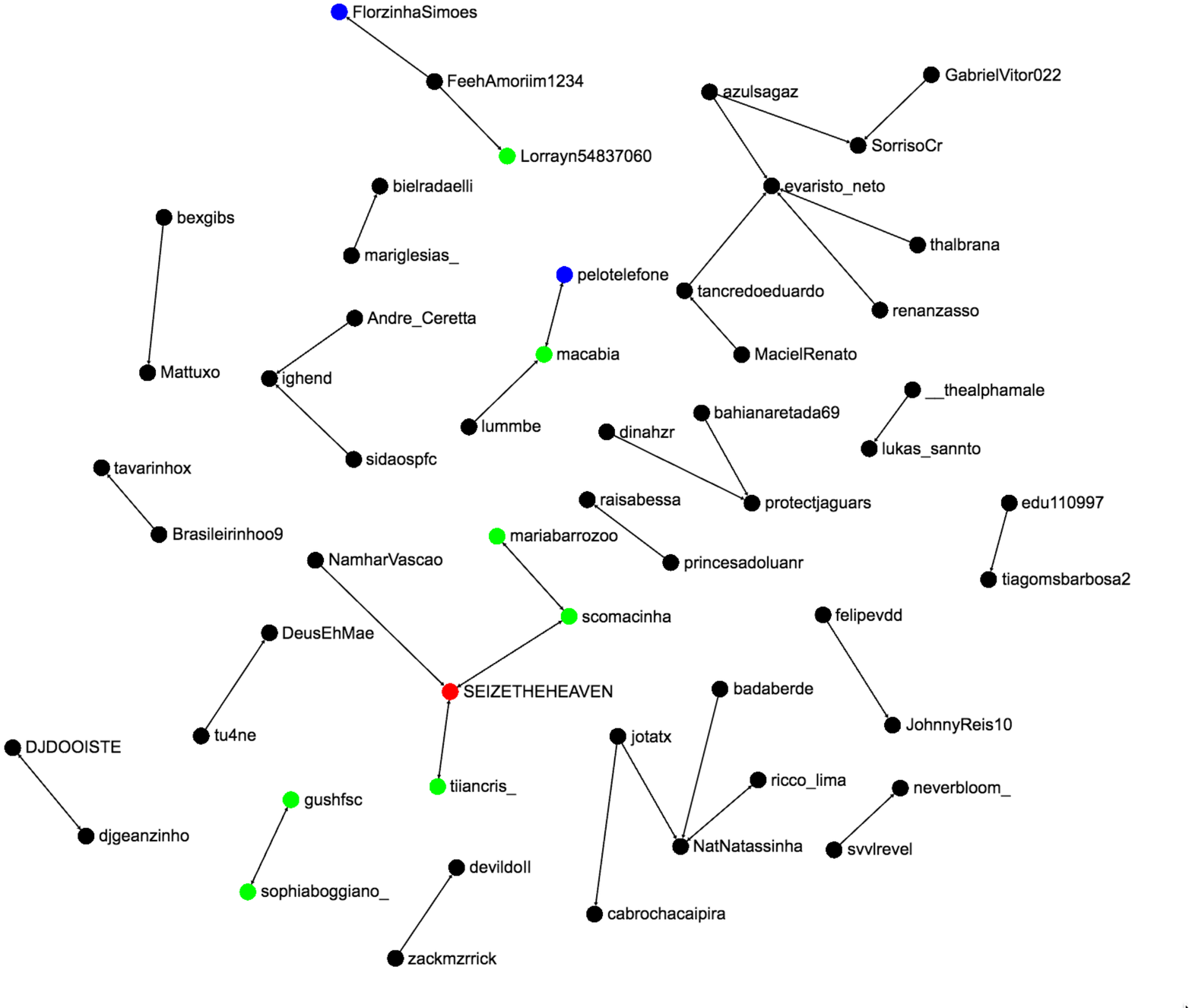}
\caption{Fragment of followers and friends graph for candidate users in our experimental dataset. Green nodes are in the top 10 TwitterRank.
Blue nodes are in top 10 TwitterRank \textit{and} top 10 Overall Focus.  Red nodes are in top 10 TwitterRank \textit{and} top 10 Topic Focus.
}
\label{fig:followers-graph}
\end{figure}

Regarding TwitterRank, we note firstly that the small absolute figures are not indicative, as the original paper \cite{weng2010twitterrank} does not provide any reference figures at all.
However we note a significant spread (150\%) between the top and bottom ranks in the top-10 list. 
The significance of this ranking, however, is questionable.
TwitterRank only yields interesting rank values when, for each user $u$, at least some of its followers are also candidate users. 
When this is not the case the approach is not very effective, because $u$'s followers' TwitterRank is a default value, which does not influence the TwitterRank of $u$ at all.
In our dataset, we find that our candidate users have very few connections amongst each other. 
This becomes clear when looking at the social connections amongst some of our candidate users, as in Fig. \ref{fig:followers-graph}.
The graph shows very promising results, as even in our small residual candidate set we discover interesting connected components, and indeed even a few friends (shown with the double arrow).
Note also that all of our top-10 TwitterRank users appears in some connected component of the graph, which is natural as it is their connectivity that contributes to their TwitterRank.
On the other hand, the number of followers of any user who actually influence the user's rank is very small.
 
We therefore compared this with the other two metrics. 
Tab. \ref{tab:TF} shows that for each of the top-10 Topic Focus users, \textit{all} of their tweets in the harvest ($T_K(u)$), however few (<10), are \texttt{Relevant}.
Furthermore, the \texttt{TF Rank} column in Tab.\ref{tab:TR} shows that all top 10 TwitterRank users are top-30 Topic Focus users, suggesting that high TwitterRankx may correlate well with high Topic Focus.

We also note that the top-10 TwitterRank user \texttt{SeizeTheHeaven} is also in the top-10 Topic Focus\footnote{User \texttt{macabia} is also in the top-10, but not shown as evidently the list of users with Topic Focus = 100 is longer than 10.}

Interestingly, if we turn to Tab. \ref{tab:OF} we see that the top-10 Overall Focus users also have a high Topic Focus, and rank within the top-20.
Again in this list we find users that rank high in other lists:  \texttt{FlorzinhaSimoes} and \texttt{pelotelefone}.

\section{Conclusions}

The research hypothesis we have explored in this paper is that social media analytics can be used to identify individuals who are actively contributing to social discourse on rthe specific topic of the Zika virus and its consequences, and are thus likely to be sensitive to health promotion campaigns. 
We tested this hypothesis by focusing on Twitter content related to the Zika virus and its effect on people.
We trained a classifier to separate the very sparse interesting signal from large amounts of noise in the feed, and then applied various ranking criteria to the set of candidate users who authored such interesting content.

Given the  sparsity of the contributors and their limited connections within the social graph, it is not surprising to find that the very popular TwitterRank algorithm \cite{weng2010twitterrank} is not particularly effective in this instance.
Despite facing a ``needle in the haystack'' problem, however, we report promising results which indicate that non topology-based metrics that count relevant tweets by user appear to be equally effective, and that a few interesting connections indeed exist in the graph amongst the top ranked users.
We are currently experimenting with larger datasets which we continually harvest from the live twitter feed.

We have developed a public-facing portal where Relevant tweets that are also geo-located are placed on a map of Brasil, and the top-k users computed using our metrics are shown and continually updated.